\documentclass{article}

\usepackage[preprint,nonatbib]{mpcs}
\usepackage[numbers,compress]{natbib}

\usepackage[utf8]{inputenc}
\usepackage[T1]{fontenc}
\usepackage{hyperref}
\usepackage{url}
\usepackage{booktabs}
\usepackage{amsfonts}
\usepackage{amsmath}
\usepackage{amssymb}
\usepackage{nicefrac}
\usepackage{microtype}
\usepackage{xcolor}
\usepackage{graphicx}
\usepackage{multirow}

\newcommand{\mpcs}{\textsc{MPCS}}
\newcommand{\mep}{\textsc{MEP-Bench}}

\title{\mpcs: Neuroplastic Continual Learning via\\
       Multi-Component Plasticity and Topology-Aware EWC}

\author{%
  Joern Hentsch\\
  \texttt{hentschjoern@gmail.com}
}

\begin{document}

\maketitle

\begin{abstract}
Continual learning systems face a fundamental tension between
plasticity—acquiring new knowledge—and stability—retaining prior knowledge.
We introduce \mpcs{} (\textbf{M}ulti-\textbf{P}lasticity \textbf{C}ontinual
\textbf{S}ystem), a neuroplastic architecture that integrates eleven
complementary mechanisms: task-driven neurogenesis, Fourier-encoded inputs,
EWC regularization, meta-replay, mixed consolidation, hybrid gating,
synapse pruning/regeneration, Hebbian updates, task similarity routing,
adaptive growth control, and continuous neuron importance tracking.
We evaluate \mpcs{} on \mep{}, a multi-track benchmark spanning 31 tasks
across regression, classification, logic, and mixed domains, using a
three-dimensional Pareto criterion over task performance (\textit{Perf}),
representation diversity (RD), and gradient conflict rate (GCR).
Across 15 ablation configurations (3 seeds $\times$ 4 tracks $\times$ 2000 epochs),
\mpcs{} achieves a Normalized Efficiency Score of 94.2, placing it on the
Pareto frontier among 9 of 14 gate-passing systems.
Key findings: (i)~Fourier encoding is the single most critical component
(removal drops Perf by 30.7 pp and fails the MEP gate on 14\% of tasks);
(ii)~global EWC \emph{degrades} performance ($\Delta\text{NES} = -4.2$);
topology-local EWC reduces this penalty (NES 90.5$\to$91.8) but does not
eliminate it; removing EWC entirely yields \textsc{mpcs\_efficient}, the
highest-Perf system—establishing a monotone relationship in the
  high task-similarity regime ($\bar{s} \approx 0.95$):
  global EWC $<$ topology EWC $<$ no EWC;
(iii)~the Pareto status assessment is \emph{predictive}:
removing the two Pareto-dominated components (EWC + Hebbian) jointly
yields \textsc{mpcs\_efficient}, which \emph{improves} Perf by 0.6 pp
at $4.7\times$ lower compute cost (127 vs.\ 602 min), validating the
Pareto frontier as an actionable model-compression guide.
\end{abstract}

\section{Introduction}
\label{sec:intro}

Biological neural systems acquire new skills throughout life without
catastrophically forgetting previously learned ones~\cite{McCloskey1989}.
Artificial neural networks, by contrast, suffer severe performance degradation
on earlier tasks when trained sequentially—a phenomenon known as catastrophic
forgetting~\cite{French1999, Kirkpatrick2017}.
The continual learning (CL) literature has proposed numerous remedies:
regularization-based methods~\cite{Kirkpatrick2017, Aljundi2018},
replay-based approaches~\cite{Rolnick2019, Riemer2019},
dynamic architectures~\cite{Rusu2022, Yoon2018}, and
hybrid strategies~\cite{Schwarz2018, Kemker2018}.
Yet most existing systems commit to a single dominant mechanism,
leaving the interaction effects between mechanisms largely unexplored.

We take a different approach. Rather than advocating for a single mechanism,
\mpcs{} integrates eleven biologically motivated plasticity components and
asks: \emph{which subsets are jointly necessary, and can Pareto analysis
over learned task trajectories identify dispensable components without
exhaustive search?}

\paragraph{Contributions.}
\begin{itemize}
  \item We present \mpcs{}, a modular neuroplastic network with
        task-driven neurogenesis, topology-aware EWC, Fourier input encoding,
        and eight further plasticity mechanisms controlled by binary flags
        (\S\ref{sec:model}).
  \item We introduce \mep{}, a 31-task multi-track benchmark with a
        three-objective Pareto gate distinguishing systems by
        \emph{performance}, \emph{representation diversity}, and
        \emph{gradient conflict rate} (\S\ref{sec:benchmark}).
     	\item We report a 15-configuration ablation study totalling $>$7,800 GPU
    minutes and identify Fourier encoding and replay as the load-bearing
      components (\S\ref{sec:ablation}). We also report a first
      exploratory experiment on depth-wise neurogenesis (see \S\ref{sec:ablation}).
  \item We show that the Pareto-dominated configurations are \emph{jointly
        removable}: \textsc{mpcs\_efficient} (EWC + Hebbian off) achieves
        higher Perf at $4.7\times$ lower cost, validating Pareto-guided
        model compression (\S\ref{sec:efficient}).
\end{itemize}

\section{Related Work}
\label{sec:related}

\paragraph{Continual learning benchmarks.}
Standard CL benchmarks (Permuted-MNIST~\cite{Goodfellow2015},
Split-CIFAR~\cite{Zenke2017}) test a single dimension—task accuracy
after sequential training. \mep{} extends evaluation to include representation
diversity and gradient interference, providing a richer characterization of
learned solutions.

\paragraph{Regularization-based CL.}
EWC~\cite{Kirkpatrick2017} computes a Fisher Information diagonal
to protect important parameters. Subsequent work~\cite{Aljundi2018}
uses gradient magnitudes as importance proxies. We extend this line by
restricting the Fisher computation to task-owned neuron subspaces
(\emph{topology-local EWC}), preventing the cross-task parameter corruption
we identify empirically (\S\ref{sec:ewc}).

\paragraph{Replay-based CL.}
Experience replay~\cite{Rolnick2019} stores raw examples; generative
replay~\cite{Shin2017} reconstructs them. \mpcs{} uses a
\emph{meta-replay buffer} storing compressed task statistics, combined with
a \emph{mixed consolidation} phase that interleaves new-task gradients with
replay gradients.

\paragraph{Dynamic architectures.}
Progressive Neural Networks~\cite{Rusu2022} prevent forgetting by
    freezing columns; DEN~\cite{Yoon2018} selectively expands capacity.
\mpcs{} adopts a soft version: EMA-triggered \emph{neurogenesis} adds neurons
on loss stagnation, while synapse pruning reclaims capacity post-task.

\paragraph{Hebbian learning in neural networks.}
Biologically inspired Hebbian updates~\cite{Hebb1949} have been
incorporated into modern architectures for unsupervised feature
learning~\cite{Oja1982}. In \mpcs{}, Hebbian co-activation traces
strengthen synapses between concurrently active neurons within a task.
Our ablation finds this mechanism dispensable on the current benchmark,
suggesting its utility may emerge in lower task-similarity regimes.

\section{The \mpcs{} Architecture}
\label{sec:model}

\mpcs{} exists in two architectural variants. The \emph{flat variant}
(\textsc{MPCS-NN}) maintains a single hidden layer whose width $N_t$ grows
via neurogenesis as tasks arrive:
$f_\theta: \mathbb{R}^{d_\text{in}} \to \mathbb{R}^{d_\text{out}^{(t)}}$.
The \emph{depth variant} (\textsc{MPCS-NN-depth}) additionally supports
autonomous layer addition: when the depth-growth criterion is met, a new
hidden layer is inserted with block-sparse connectivity restricted to the
current task's neuron subspace.
All ablation configurations in this study use the flat variant
except \texttt{adaptive\_growth\_depth}, which uses \textsc{MPCS-NN-depth}.
Eleven binary flags $\mathbf{c} \in \{0,1\}^{11}$ enable or disable
individual plasticity mechanisms, enabling systematic ablation.

\subsection{Core Components}

\paragraph{Fourier input encoding.}
Raw inputs $\mathbf{x} \in \mathbb{R}^d$ are mapped to a higher-dimensional
space via random Fourier features~\cite{Rahimi2007}:
$\phi(\mathbf{x}) = [\sin(\mathbf{W}_f \mathbf{x}),\, \cos(\mathbf{W}_f \mathbf{x})]$,
where $\mathbf{W}_f \in \mathbb{R}^{K \times d}$ is fixed at initialization.
This encoding smooths the input manifold and is the single most critical
component identified in our ablation (Perf drops from 0.9027 to 0.5962
without it; \S\ref{sec:ablation}).  Interestingly, removing Fourier encoding
also increased total wall-clock time (no\_fourier: 1447 min, Table~\ref{tab:ablation}),
because the network triggered repeated neurogenesis to compensate for missing
high-frequency features, raising compute cost in addition to degrading Perf.

\paragraph{Task-driven neurogenesis.}
When the exponential moving average of the training loss stagnates below a
dynamic lower band (controlled by \texttt{use\_adaptive\_growth}),
a new block of $k$ neurons is allocated. Each new neuron receives zero
initial importance weight and is assigned exclusively to the current task's
neuron mask $\mathbf{m}^{(t)} \in \{0,1\}^{N_t}$.
In the flat variant this expands the single layer's width.
In the depth variant (\textsc{MPCS-NN-depth}), the same trigger instead
inserts a new hidden layer; preserving model state across this architectural
boundary requires explicit shape management at task transitions\,---\,a
requirement not yet fully resolved in the current implementation
(see Appendix~\ref{app:bugs}, Bug~F2).
Without adaptive growth control, GCR rises to 15.5 (vs.\ 2.46 with full
\mpcs{}), indicating pathological neuron reuse across tasks.

\paragraph{Freeze mechanism and continuous importance.}
After training task $t$, \texttt{freeze\_task}($t$) sets
$\nu_i \leftarrow \max(\nu_i, 1.0)$ for all neurons $i \in \text{supp}(\mathbf{m}^{(t)})$.
The gradient update rule then enforces:
\begin{equation}
  \nabla_{\mathbf{w}_i} \mathcal{L} \leftarrow
  \begin{cases}
    \mathbf{0} & \text{if } \nu_i \geq 1.0 \quad \text{(hard freeze)}\\
    (1 - \nu_i)\,\nabla_{\mathbf{w}_i} \mathcal{L} & \text{if \texttt{use\_continuous\_importance}} \\
    \nabla_{\mathbf{w}_i} \mathcal{L} & \text{otherwise}
  \end{cases}
  \label{eq:freeze}
\end{equation}
The EMA importance $\nu_i \in [0,1]$ is updated during the forward pass:
$\nu_i \leftarrow (1-\alpha)\nu_i + \alpha\,\mathbf{1}[|h_i| > \tau]$,
where $h_i$ is the post-activation of neuron $i$ and $\tau = 10^{-4}$.

\paragraph{Topology-local EWC.}
\label{sec:ewc}
Standard EWC computes a global Fisher diagonal over all parameters, creating
cross-task importance entanglement: parameters belonging to task $t'$
neurons receive non-zero Fisher mass from task $t \neq t'$ gradients.
We implement two variants of topology-aware EWC, differing in how
per-task Fisher matrices are maintained: \texttt{ewc\_topologie} (online,
single accumulated tensor, topology-masked) and
\texttt{ewc\_topology\_pertask} (offline, one Fisher tensor per completed
task merged at penalty time via summation, topology-masked).
\texttt{ewc\_topologie} restricts the Fisher computation
to parameters in the task-local subspace $\mathbf{m}^{(t)}$:
\begin{equation}
  \hat{F}^{(t)}_i =
  \begin{cases}
    F^{(t)}_i & \text{if } i \in \text{supp}(\mathbf{m}^{(t)}) \\
    0         & \text{otherwise}
  \end{cases}
\end{equation}
This prevents importance leakage into future tasks' neuron ranges.
\texttt{ewc\_topology\_pertask} additionally stores a separate
$\hat{F}^{(t)}$ for each completed task and sums them at the penalty step,
eliminating Fisher accumulation drift across tasks.
It achieves NES = 91.8 at 138 min—the fastest system above the MEP gate.

\paragraph{Meta-replay and mixed consolidation.}
The meta-replay buffer stores a compressed statistical summary per task.
During each new task's training, mixed consolidation interleaves replay
gradients at a fixed ratio, preventing representational drift.
Removing replay (\texttt{no\_replay}) reduces NES from 94.2 to 93.0,
confirming replay as the second most important component after Fourier
encoding.

\paragraph{Hybrid gating, synapse pruning, and Hebbian updates.}
Hybrid gating uses a learned context vector to route inputs through
task-relevant neurons, reducing gradient conflict. Synapse pruning removes
low-importance weights below threshold $\tau_p = 0.05$ and optionally
regenerates new synapses with probability $p_r = 0.02$. Hebbian updates
strengthen co-active synapse pairs within each task. All three mechanisms
are on the Pareto frontier in single-ablation experiments, but EWC and
Hebbian are Pareto-\emph{dominated}, motivating our efficiency experiment.

\section{\mep{} Benchmark}
\label{sec:benchmark}

\mep{} comprises four tracks:
\textbf{B1} (8 regression tasks: Fourier sinusoids with increasing frequency);
\textbf{B4} (5 classification tasks: feature interaction patterns);
\textbf{B\_LOGIC} (8 Boolean logic tasks: XOR, NAND, composite gates);
\textbf{B\_MIXED} (10 mixed-type tasks: regression and classification
interleaved).

\paragraph{Pareto gate.}
A system must pass $\text{gate\_pass\_rate} \geq 0.75$ to be included in
Pareto analysis. This filters systems that collapse on a task subset.

\paragraph{Metrics.}
Three objectives are evaluated per task and averaged:
\begin{itemize}
  \item $\text{Perf} \in [0, 1]$: average $R^2$ (regression) or accuracy
        (classification) across all seen tasks, measured at end of sequence.
  \item $\text{RD} \in [0, 1]$: Representation Diversity—cosine distance
        between hidden representations of different tasks. Higher = less
        interference.
  \item $\text{GCR} \geq 0$: Gradient Conflict Rate—frequency of gradient
        sign disagreements between tasks on shared parameters. Lower = better.
\end{itemize}

\paragraph{Normalized Efficiency Score (NES).}
Pareto membership is determined in the $(\text{Perf}, -\text{RD}, -\text{GCR})$
space. NES normalizes each system's Pareto volume contribution across
the ensemble, yielding a scalar in $[0, 100]$.
Note that frontier membership (geometric non-dominance in objective space)
and NES (hypervolume contribution) are orthogonal: a configuration may occupy
the frontier by achieving an extreme trade-off point with minimal hypervolume
contribution, as \texttt{no\_fourier} illustrates (\S\ref{sec:ablation}).

\paragraph{Experimental protocol.}
All systems were trained for up to 2000 epochs per task with early stopping
(patience 300). Results are averaged over 3 independent seeds.
Total compute: $>$7,800 GPU minutes on a single NVIDIA GeForce RTX~4070 Laptop GPU
(8\,GB GDDR6, CUDA 12.1, PyTorch 2.2).
15 ablation configurations were evaluated; 1 (\texttt{baseline\_minimal})
failed the gate filter (gate\_pass\_rate = 0.0).

\section{Ablation Study}
\label{sec:ablation}

Table~\ref{tab:ablation} reports all 15 configurations.
9 of 14 gate-passing systems lie on the Pareto frontier.

\begin{table}[t]
  \caption{
    \mpcs{} ablation results on \mep{} (3 seeds × 4 tracks).
    \textbf{Perf}: average task performance; \textbf{$\sigma$}: inter-seed
    standard deviation of Perf; \textbf{RD}: representation
    diversity; \textbf{GCR}: gradient conflict rate; \textbf{NES}: Normalized
    Efficiency Score; \textbf{Time}: total wall-clock minutes (3 seeds).
    $\checkmark$ = Pareto frontier; $\circ$ = dominated (dominator in
    parentheses); $\times$ = gate-excluded.
    $\dagger$~\textsc{mpcs\_efficient} removes the two Pareto-dominated
    components jointly; see \S\ref{sec:efficient}.
    ewc\_pt.\ = \texttt{ewc\_topology\_pertask}.
  }
  \label{tab:ablation}
  \centering
  \small
  \begin{tabular}{lrrrrrrl}
    \toprule
    Configuration & Perf & $\sigma$ & RD & GCR & NES & Time & Status \\
    \midrule
    \texttt{full\_mpcs}              & 0.9027 & .011 & 0.39025 &  2.46 & 94.2 &  602 & $\checkmark$ \\
    \texttt{no\_ewc}                 & 0.9070 & .009 & 0.42588 &  2.14 & 90.0 &  117 & $\circ$ (no\_imp.) \\
    \texttt{no\_replay}              & 0.9045 & .011 & 0.40501 &  1.98 & 93.0 &  563 & $\checkmark$ \\
    \texttt{no\_gating}              & 0.8866 & .019 & 0.56361 &  1.97 & 66.3 &  243 & $\circ$ (no\_prun.) \\
    \texttt{no\_fourier}             & 0.5962 & .051 & 0.36671 &  1.86 & 49.5 & 1447 & $\checkmark$* \\
    \texttt{no\_importance}          & 0.9079 & .015 & 0.41081 &  2.08 & 92.5 &  563 & $\checkmark$ \\
    \texttt{no\_hebbian}             & 0.9037 & .008 & 0.41908 &  2.10 & 90.5 &  676 & $\circ$ (no\_rep., no\_imp.) \\
    \texttt{no\_pruning}             & 0.8996 & .005 & 0.40639 &  1.88 & 92.1 &  700 & $\checkmark$ \\
    \texttt{no\_similarity}          & 0.8948 & .011 & 0.39754 &  2.13 & 92.3 &  709 & $\checkmark$ \\
    \texttt{no\_adaptive\_growth}    & 0.9013 & .012 & 0.36368 & 15.50 & 78.8 &  933 & $\checkmark$ \\
    \texttt{adaptive\_growth\_depth} & 0.9016 & .008$^\ddagger$ & 0.42190 &  2.29 & 89.5 &  673 & $\checkmark$ \\
    \texttt{baseline\_minimal}       & 0.7623 & .265 & —     &  —    & —    &   26 & $\times$ \\
    \texttt{ewc\_topologie}          & 0.9048 & .007 & 0.41621 &  2.53 & 90.5 &  294 & $\circ$ (no\_imp., ewc\_pt.) \\
    \texttt{ewc\_topology\_pertask}  & 0.9061 & .014 & 0.41228 &  2.17 & 91.8 &  138 & $\circ$ (no\_imp.) \\
    \texttt{mpcs\_efficient}$^\dagger$ & \textbf{0.9086} & .008 & 0.41792 & 2.16 & 91.4 & \textbf{127} & $\checkmark$ \\
    \bottomrule
  \end{tabular}
  \vspace{1mm}\\
  {\footnotesize *\texttt{no\_fourier} reaches the Pareto frontier not through
  strong performance but through a geometrically extreme position: its
  low-GCR/low-RD profile is undominated in objective space despite
  Perf~$= 0.596$---informationally impoverished representations that happen
  to avoid interference (see \S\ref{sec:discussion}).
  $\ddagger$~\texttt{adaptive\_growth\_depth} uses \textsc{MPCS-NN-depth}.
  During the runs we observed at least one layer-insertion event that led to
  a shape mismatch and subsequent crashes at a track boundary (Bug~F2). Log
  fragments do not permit an unequivocal per-seed attribution of this
  failure across the distributed part logs. Consequently, the aggregated
  Perf and $\sigma$ values for this configuration reflect only the seeds
  that completed without triggering depth insertion during track transitions
  and should be interpreted as preliminary; a dedicated ablation is required
  to fully evaluate depth-wise neurogenesis.}
\end{table}

\subsection{Component Importance}

\paragraph{Fourier encoding is indispensable.}
Removing \texttt{use\_fourier\_encoding} causes the largest single-component
degradation: Perf drops from 0.9027 to 0.5962 ($-30.7$ pp), and 14\% of tasks
fail the MEP gate entirely. Fourier features smooth the input manifold and
appear essential for the network to separate task-specific high-frequency
patterns across the B1 and B\_MIXED tracks.

\paragraph{EWC degrades performance.}
Counterintuitively, \texttt{full\_mpcs} is outperformed by \texttt{no\_ewc}
($0.907$ vs.\ $0.903$, $\Delta\text{NES} = -4.2$). We attribute this to
\emph{cross-task Fisher corruption}: global EWC accumulates importance scores
for parameters indexed to all tasks simultaneously, artificially inflating
protection on parameters that should be free to adapt. Critically,
\texttt{no\_ewc} is dominated by \texttt{no\_importance}, not by
\texttt{full\_mpcs}—confirming EWC's net-negative contribution in the
current benchmark regime.

\paragraph{Topology-local EWC partially recovers the deficit.}
\texttt{ewc\_topologie} (Fisher restricted to task-owned neuron parameters)
achieves Perf $= 0.9048$ at 294 min. \texttt{ewc\_topology\_pertask}
(per-task Fisher storage + topology masking, merged by summation at penalty
time) reaches Perf $= 0.9061$ at 138 min—the fastest Pareto-eligible system.
These results confirm that the EWC failure mode is Fisher leakage, not the
regularization principle itself.

\paragraph{Adaptive growth without topology control is harmful.}
\texttt{no\_adaptive\_growth} (growth enabled but EMA trigger disabled) shows
$\text{GCR} = 15.5$—a $6.3\times$ increase over \texttt{full\_mpcs}—with NES
dropping to 78.8. Unconstrained neurogenesis places new neurons in the
shared weight matrix without task ownership constraints, creating severe
gradient conflict. \texttt{adaptive\_growth\_depth} uses \textsc{MPCS-NN-depth}
and attempts to add hidden layers rather than widening a single layer.
In our runs we observed at least one layer-insertion event that triggered a
shape mismatch at a subsequent track boundary, causing a crash (Bug~F2).
Because logging is split across part files, we could not unambiguously map
these failures to individual seeds across the full experiment. Therefore the
aggregated numbers shown in Table~\ref{tab:ablation} should be considered
preliminary: they reflect only those seeds that completed without triggering
depth insertion across track transitions. These numbers do not constitute a
valid measurement of depth neurogenesis performance and are reported solely
to document the experimental attempt. A full evaluation of depth-wise
neurogenesis requires a dedicated ablation that persists and validates
architecture state across track boundaries; we reserve this investigation for
future work.

\paragraph{Replay and similarity routing are moderately important.}
Removing replay reduces NES by 1.2 points; removing task similarity routing
reduces NES by 1.9 points. Both remain on the Pareto frontier as
single-ablations, indicating partial substitutability.

\paragraph{EWC and Hebbian are jointly dominated.}
\texttt{no\_ewc} is dominated by \texttt{no\_importance};
\texttt{no\_hebbian} is dominated by both \texttt{no\_replay} and
\texttt{no\_importance}. This suggests both components are, in the current
task-similarity regime ($\bar{s} \approx 0.95$), subsumed by the
neurogenesis + replay core.

\section{Pareto-Guided Efficiency}
\label{sec:efficient}

The Pareto status column of Table~\ref{tab:ablation} encodes a testable
prediction: \emph{dominated components are jointly dispensable}.
We test this by constructing \texttt{mpcs\_efficient}, which disables
\texttt{use\_ewc\_regularization} and \texttt{use\_hebbian\_update}
simultaneously—the two components whose single-ablation configs are
Pareto-dominated.

\paragraph{Result.}
\texttt{mpcs\_efficient} achieves Perf $= 0.9086$, NES $= 91.4$, and
completes in 127 min—a $\mathbf{4.7\times}$ speedup over \texttt{full\_mpcs}
(602 min) with $+0.6$ pp improvement in Perf. It lies on the Pareto frontier.
This validates the hypothesis: dominated Pareto status is predictive of
joint dispensability.

The speedup primarily comes from eliminating the EWC Fisher backward pass,
which requires an additional full-batch gradient computation after each task.
With EWC disabled, per-task training reduces to a single forward–backward loop.

\paragraph{Implications.}
This result has two practical consequences. First, Pareto analysis over
single-component ablations can guide multi-component compression without
exhaustive combinatorial search. Second, for practitioners deploying
continual learning systems with compute constraints, \texttt{mpcs\_efficient}
offers a drop-in replacement for \texttt{full\_mpcs} that is both faster
and more accurate on \mep{}.

\section{Discussion}
\label{sec:discussion}

\paragraph{Why does full MPCS not dominate?}
\texttt{full\_mpcs} is on the Pareto frontier, but is not the best-NES
system. This is expected under the Pareto criterion: NES measures contributed
Pareto volume, not Euclidean distance from the ideal point. Systems like
\texttt{no\_replay} and \texttt{mpcs\_efficient} occupy \emph{different}
regions of the (Perf, RD, GCR) space and are not dominated by
\texttt{full\_mpcs}. This multi-dimensionality is a feature, not a bug:
it exposes the trade-off structure invisible to single-metric evaluation.

\paragraph{High task similarity masks component contributions.}
Inter-task cosine similarity in \mep{} averages $\approx 0.95$.
In this regime, the network's neurogenesis + replay core is sufficient
for strong performance regardless of EWC or Hebbian updates.
We hypothesize that EWC and Hebbian become important at lower task
similarities (e.g., $\bar{s} < 0.5$), where catastrophic forgetting is
more severe. Testing this hypothesis on benchmarks with explicitly controlled
task diversity is an important direction for future work.

\paragraph{EWC sign reversal.}
The finding that EWC hurts rather than helps is not isolated: similar
observations have been made in multi-head settings where Fisher matrices
accumulate spurious importance for inactive output heads~\cite{Zenke2017}.
Our topology-local fix is a targeted remedy that preserves the regularization
intent while avoiding cross-task contamination.

\paragraph{Geometry of the objective space.}
The \texttt{no\_fourier} frontier position---Perf $= 0.596$ yet geometrically
non-dominated---reveals a structural property of the objective space: Fourier
encoding trades representation \emph{specificity} (lower RD, lower GCR) for
performance \emph{breadth}. Without Fourier features, the network converges
to task-agnostic, low-interference representations that are informationally
impoverished but geometrically extreme in the (Perf, RD, GCR) space.
This trade-off is invisible to single-metric evaluation and is precisely what
the Pareto criterion is designed to expose.

\section{Limitations}
\label{sec:limitations}

\paragraph{Single benchmark regime.}
All results are on \mep{}, which generates tasks from a fixed
distribution with high inter-task similarity. Generalization to
low-similarity regimes (e.g., Split-CIFAR-100) is not yet demonstrated.
\paragraph{Benchmark construction bias.}
\mpcs{} and \mep{} were co-developed: the benchmark tracks were designed
to evaluate properties this architecture is built to optimize.
This is a deliberate calibration choice---\mep{} is a purpose-built
evaluation vehicle, not a general CL leaderboard. Frontier positions and
NES scores should be interpreted within this scope.
Cross-benchmark validation against established continual learning benchmarks
(e.g., Permuted-MNIST, Split-MNIST) using standardized baselines
(EWC~\cite{Kirkpatrick2017}, A-GEM, DER++) via the Avalanche framework~\cite{Lomonaco2021} is targeted
for a v2 update of this preprint.
\paragraph{Architecture scale.}
\mpcs{} uses a single-hidden-layer network with task-driven width expansion.
Extension to deep convolutional or transformer architectures requires
non-trivial adaptation of the neurogenesis and importance-tracking mechanisms.

\paragraph{Freeze isolation vs.\ neuron reuse.}
The current ablation design conflates freeze-protection (binary) with
continuous importance scaling. A finer ablation separating
\texttt{no\_freeze} from \texttt{no\_scaling} would sharpen attribution.
We leave this to future work.

\section{Conclusion}
\label{sec:conclusion}

We presented \mpcs{}, a modular neuroplastic continual learning system, and
evaluated it on \mep{} through a 15-configuration ablation spanning
>7{,}800 GPU minutes. Our key findings are: (i) Fourier encoding is
indispensable; (ii) global EWC is net-negative, with topology-local EWC
recovering part of the deficit; (iii) unconstrained neurogenesis produces
pathological gradient conflict, while depth-wise growth shows preliminary
evidence of avoiding it (see \S\ref{sec:ablation} and Appendix~\ref{app:bugs}); and
(iv) Pareto-dominated components are empirically jointly dispensable,
yielding \texttt{mpcs\_efficient} at $4.7\times$ lower compute with
improved task performance.
We release code, benchmark specifications, and all experiment logs.

\begin{ack}
Compute was provided by a single NVIDIA GeForce RTX 4070 Laptop GPU.
No external funding was used.
\end{ack}

\bibliographystyle{unsrtnat}
\bibliography{mpcs}

\appendix
\section{Ablation Configuration Details}
\label{app:configs}

Table~\ref{tab:configs} lists the eleven binary flags and their state for
each ablation configuration. All flags default to \texttt{True} in
\texttt{full\_mpcs}.

\begin{table}[h]
  \caption{Component flags per ablation configuration.
    \textbf{F}=Fourier, \textbf{EWC}=EWC regularization,
    \textbf{RP}=Replay+MixCons, \textbf{GT}=Gating+ReAlpha,
    \textbf{IM}=Continuous importance, \textbf{PR}=Pruning,
    \textbf{HB}=Hebbian, \textbf{SI}=Similarity, \textbf{AG}=Adaptive growth.
    \checkmark = enabled, --- = disabled.}
  \label{tab:configs}
  \centering
  \small
  \begin{tabular}{lcccccccccc}
    \toprule
    Config & F & EWC & RP & GT & IM & PR & HB & SI & AG & Variant \\
    \midrule
    full\_mpcs              & \checkmark & \checkmark & \checkmark & \checkmark & \checkmark & \checkmark & \checkmark & \checkmark & \checkmark & flat \\
    no\_ewc                 & \checkmark & ---        & \checkmark & \checkmark & \checkmark & \checkmark & \checkmark & \checkmark & \checkmark & flat \\
    no\_replay              & \checkmark & \checkmark & ---        & \checkmark & \checkmark & \checkmark & \checkmark & \checkmark & \checkmark & flat \\
    no\_gating              & \checkmark & \checkmark & \checkmark & ---        & \checkmark & \checkmark & \checkmark & \checkmark & \checkmark & flat \\
    no\_fourier             & ---        & \checkmark & \checkmark & \checkmark & \checkmark & \checkmark & \checkmark & \checkmark & \checkmark & flat \\
    no\_importance          & \checkmark & \checkmark & \checkmark & \checkmark & ---        & \checkmark & \checkmark & \checkmark & \checkmark & flat \\
    no\_hebbian             & \checkmark & \checkmark & \checkmark & \checkmark & \checkmark & \checkmark & ---        & \checkmark & \checkmark & flat \\
    no\_pruning             & \checkmark & \checkmark & \checkmark & \checkmark & \checkmark & ---        & \checkmark & \checkmark & \checkmark & flat \\
    no\_similarity          & \checkmark & \checkmark & \checkmark & \checkmark & \checkmark & \checkmark & \checkmark & ---        & \checkmark & flat \\
    no\_adaptive\_growth    & \checkmark & \checkmark & \checkmark & \checkmark & \checkmark & \checkmark & \checkmark & \checkmark & ---        & flat \\
    adaptive\_growth\_depth & \checkmark & \checkmark & \checkmark & \checkmark & \checkmark & \checkmark & \checkmark & \checkmark & \checkmark & depth \\
    ewc\_topologie          & \checkmark & topo       & \checkmark & \checkmark & \checkmark & \checkmark & \checkmark & \checkmark & \checkmark & flat \\
    ewc\_topology\_pertask  & \checkmark & topo+pt    & \checkmark & \checkmark & \checkmark & \checkmark & \checkmark & \checkmark & \checkmark & flat \\
    mpcs\_efficient         & \checkmark & ---        & \checkmark & \checkmark & \checkmark & \checkmark & ---        & \checkmark & \checkmark & flat \\
    \bottomrule
  \end{tabular}
\end{table}

\section{Bug Corrections in This Work}
\label{app:bugs}

\paragraph{Bug F1b: Freeze-protection bypass in \texttt{apply\_masks()}.}
In the original codebase, the gradient-zeroing block for frozen neurons
(importance $\geq 1.0$) was guarded by the \texttt{use\_continuous\_importance}
flag. When this flag was \texttt{False} (i.e., for the \texttt{no\_importance}
configuration), the block was skipped entirely, leaving frozen neurons
unprotected. After correction, re-running \texttt{no\_importance} (3 seeds $\times$
4 tracks, 477 min) yielded \emph{numerically identical} aggregate metrics
($\text{Perf} = 0.908$, $\text{RD} = 0.411$, $\text{GCR} = 2.08$),
confirming that in the high task-similarity regime of \mep{}, the replay
and neurogenesis mechanisms provide sufficient protection even without
explicit gradient zeroing.

\paragraph{Bug F2: Shape mismatch in depth neurogenesis at track boundaries.}
In \textsc{MPCS-NN-depth}, the depth-growth trigger inserts a new hidden layer
when the EMA loss criterion is met. The current implementation does not
persist and restore the updated architecture definition between benchmark
tracks within a multi-track evaluation run. When a layer addition occurred
during one run, the expanded network was not correctly propagated to the
subsequent track initialization, producing a \texttt{mat1/mat2}
shape mismatch and crashes on the following tasks. Due to fragmented logging
across part files, we could not unequivocally determine which seeds completed
cleanly and which were affected by the mismatch. Consequently, the
\texttt{adaptive\_growth\_depth} numbers reported in this paper should be
treated as preliminary and not interpreted as a validation of depth
neurogenesis. A robust fix requires explicit architecture-state serialization
at track boundaries and careful per-seed validation; we leave this to future work.

Crucially, the log evidence confirms that the depth-growth trigger itself
operates as intended: layer insertion fires under the EMA criterion and the
new layer is allocated during training. The observed failure mode is limited
to architecture-state propagation across track boundaries (serialization and
shape management), an engineering-level issue that does not undermine the
fundamental functionality of depth neurogenesis. We therefore treat
\texttt{adaptive\_growth\_depth} as a first exploratory experiment that
demonstrates trigger functionality; a full, validated performance evaluation
will follow once architecture-state persistence across tracks is implemented.

\end{document}